\documentclass[10pt,twocolumn,letterpaper]{article}

\usepackage{cvpr}
\usepackage{times}
\usepackage{epsfig}
\usepackage{graphicx}
\usepackage{amsmath}
\usepackage{amssymb}
\usepackage{gensymb}
\usepackage{textcomp}
\usepackage{multirow}
\usepackage{mathtools,xparse}
\usepackage{enumitem}
\usepackage{booktabs}
\usepackage{verbatim}
\usepackage{microtype} 
\usepackage{array}
\usepackage{floatrow}

\usepackage[pagebackref=true,breaklinks=true,colorlinks,bookmarks=false]{hyperref}

\cvprfinalcopy 

\def\cvprPaperID{4363} 

\ifcvprfinal\fi
\begin{document}

\title{\vspace{-2em}Deep Sky Modeling for Single Image Outdoor Lighting Estimation}

\author{Yannick Hold-Geoffroy$^*$\\
Adobe Research\\
{\tt\small holdgeof@adobe.com}
\and
Akshaya Athawale\thanks{Parts of this work were completed while Y. Hold-Geoffroy and A. Athawale were at U. Laval.}\\
Indian Institute of Tech. Dhanbad\\
{\tt\small akshaya.15je001564@am.ism.ac.in}
\and
Jean-Fran\c{c}ois Lalonde\\
Universit\'{e} Laval\\
{\tt\small jflalonde@gel.ulaval.ca}
\vspace{-4em}
}

\maketitle

\begin{abstract}
We propose a data-driven learned sky model, which we use for outdoor lighting estimation from a single image. As no large-scale dataset of images and their corresponding ground truth illumination is readily available, we use complementary datasets to train our approach, combining the vast diversity of illumination conditions of SUN360 with the radiometrically calibrated and physically accurate Laval HDR sky database. Our key contribution is to provide a holistic view of both lighting modeling and estimation, solving both problems end-to-end. From a test image, our method can directly estimate an HDR environment map of the lighting without relying on analytical lighting models. We demonstrate the versatility and expressivity of our learned sky model and show that it can be used to recover plausible illumination, leading to visually pleasant virtual object insertions. To further evaluate our method, we capture a dataset of HDR 360$^\circ$ panoramas and show through extensive validation that we significantly outperform previous state-of-the-art. 
\end{abstract}

\vspace{-1em}
\section{Introduction}
\label{sec:introduction}

The lighting conditions of outdoor scenes can create significant differences in the scene appearance depending on the weather and the time of day. Indeed, one need only consider the striking contrast created by bright highlights and dark shadows at noon, the warm, orange hues of the golden hour, or the gray ominous look of overcast conditions. This wide variety of effects is challenging for approaches that attempt to estimate the lighting conditions from outdoor images. 

\begin{figure}[t]
\centering
\includegraphics[width=0.8\linewidth]{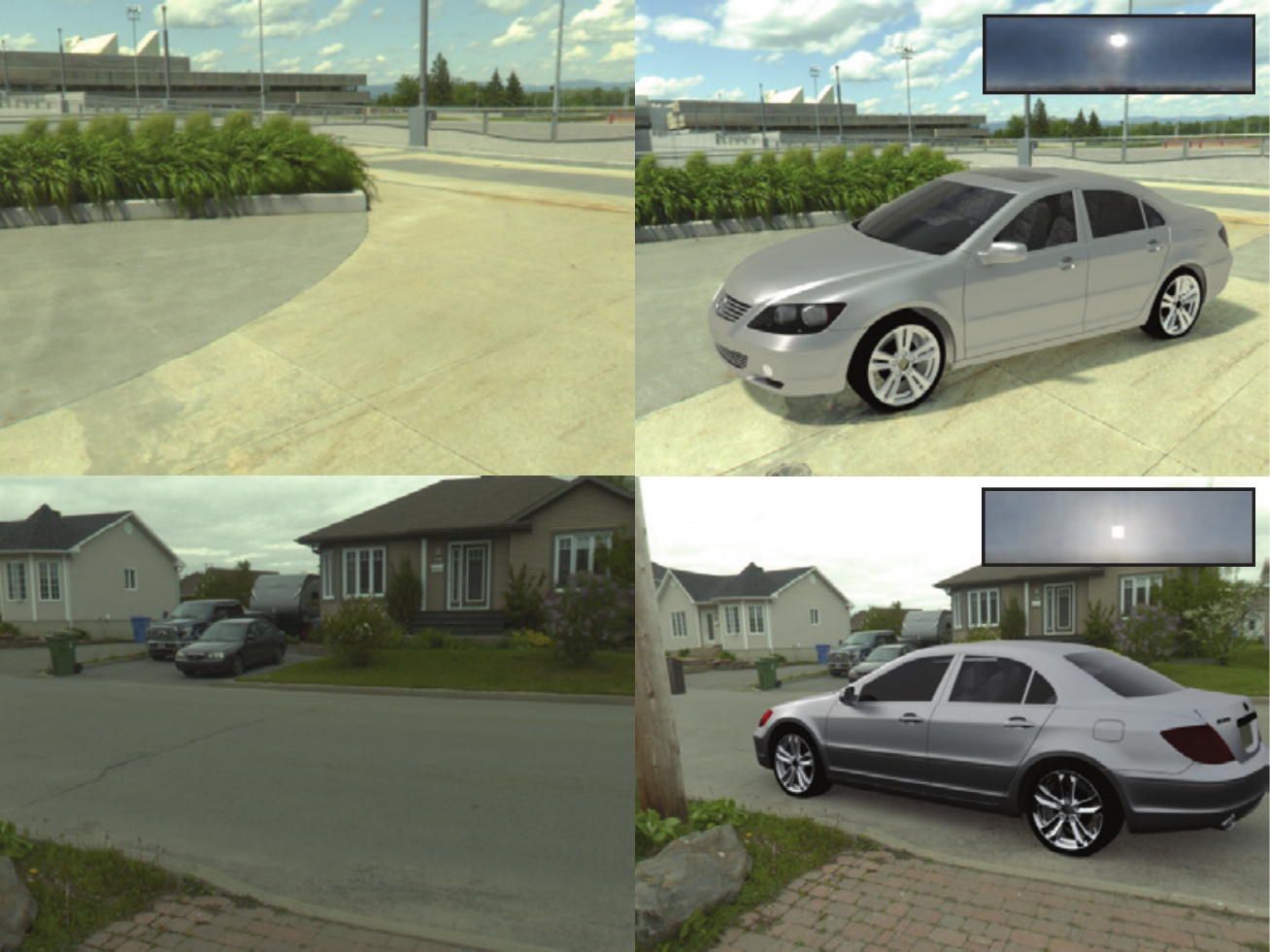}
\caption{Our method can estimate HDR outdoor lighting conditions from a single image (left). This estimation can be used ``as-is'' to relight virtual objects that match the input image in both sunny (top-right) and overcast (bottom-right) weather. Our key contribution is to train both our sky model and lighting estimation end-to-end by exploiting multiple complementary datasets during training. \vspace{-1.2em}}
\label{fig:teaser}
\end{figure}

A popular solution to this problem involves capturing objects of known geometry and reflectance properties (notably, a chrome sphere~\cite{Debevec1998}). Another solution, which does not require access to the scene, is to approximate outdoor lighting with low-dimensional, parametric models. This has the advantage of drastically reducing the dimensionality of the problem down to just a handful of variables, which can more easily be estimated from an image. This insight has recently been exploited to successfully learn to predict lighting from a single outdoor image~\cite{holdgeoffroy-cvpr-17}. In particular, they propose to represent outdoor lighting using the Ho\v{s}ek-Wilkie (HW) sky model~\cite{hosek-siggraph-12,hosek-cga-13}, which can model high dynamic range (HDR) sky domes using as few as 4 parameters. They learn to predict lighting by fitting the HW model to a large database of outdoor, low dynamic range (LDR) panoramas and training a CNN to regress the HW parameters from limited field of view crops extracted from those panoramas. 

\begin{figure}
\centering
\newcolumntype{C}[1]{>{\centering\let\newline\\\arraybackslash\hspace{0pt}}m{#1}}
\setlength{\tabcolsep}{0pt}
\begin{tabular}{C{2.15cm}C{2.15cm}C{2.15cm}}
Ground truth & Ours & \cite{hosek-siggraph-12,hosek-cga-13} \\
\end{tabular}
\includegraphics[width=0.8\linewidth]{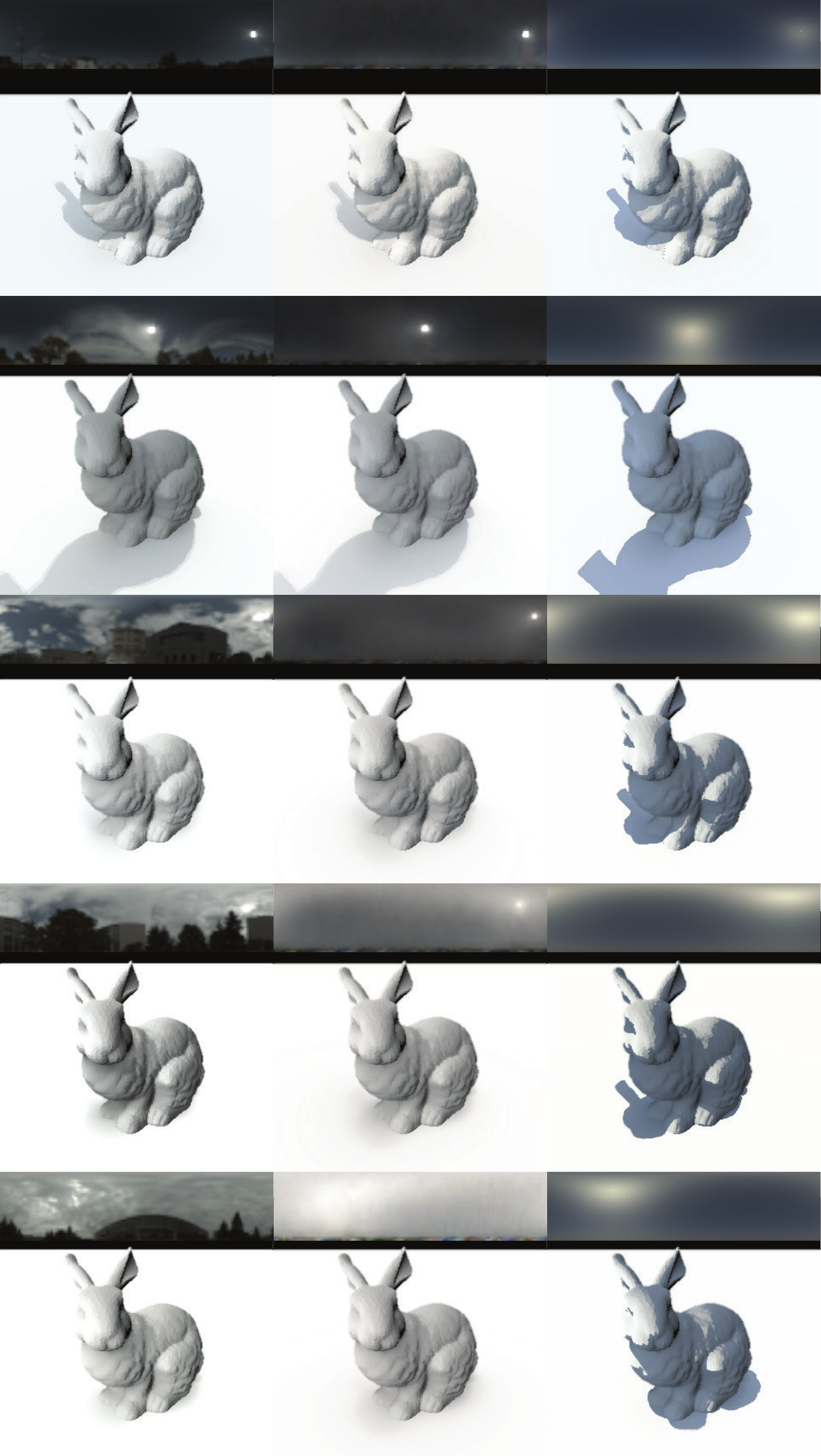}
\caption{Examples of our 360\textdegree\ unsaturated HDR database (left), our reconstruction using our learned sky model (center) and the Ho\v{s}ek-Wilkie sun and sky models~\cite{hosek-siggraph-12,hosek-cga-13} fit using the optimization described in~\cite{holdgeoffroy-cvpr-17} (right). Renders of each method are shown below the panorama. Note how our sky model can accurately produce a wide variety of lighting conditions from sunny (top) to overcast (bottom) and their corresponding shadow contrast and smoothness. \vspace{-1em}}
\label{fig:dataset-excerpt}
\end{figure}

Unfortunately, approximating outdoor lighting analytically comes at a cost. Popular sky models (e.g. \cite{hosek-siggraph-12,preetham-siggraph-99,perez1993allweather}) were developed to model \emph{clear} skies with smoothly-varying amounts of atmospheric aerosols (represented by the commonly-used \emph{turbidity} parameter). Therefore, they do not yield accurate representations for other types of common weather conditions such as partially cloudy or completely overcast skies. For example, consider the different lighting conditions in fig.~\ref{fig:dataset-excerpt}, which we represent with the HW parametric model using the non-linear fitting approach of~\cite{holdgeoffroy-cvpr-17}. Note how the HW approximation works well in clear skies (top) but degrades as the cloud cover increases (bottom). Can we obtain a lighting model that is low-dimensional, that can accurately describe the wide variety of outdoor lighting conditions, and that can be estimated from a single image? 

In this paper, we propose an answer to this question by \emph{learning} an HDR sky model directly from data. Our non-analytical data-driven sky model can be estimated directly from a single image captured outdoors. Our approach successfully models a much larger set of lighting conditions than previous approaches (see fig.~\ref{fig:dataset-excerpt}). 

To learn to estimate a non-parametric lighting model from a single photograph, we propose a three-step approach which bears resemblance to the ``T-network'' architecture proposed by \cite{girdhar2016learning}, and rely on a variety of existing complementary datasets. First, we train a deep sky autoencoder that learns a data-driven, deep HDR sky model. To train this sky autoencoder, we rely on the Laval HDR sky database~\cite{hdrdb,lalonde-3dv-14}, a large dataset of unsaturated HDR hemispherical sky images. Second, we project the SUN360 LDR outdoor panorama dataset~\cite{xiao-cvpr-12} to HDR, using the ``LDR2HDR'' network of Zhang and Lalonde~\cite{zhang-iccv-2017}, and subsequently map each panorama to the latent space of HDR skies from our sky autoencoder. This effectively provides non-parametric sky labels for each panorama. Third, we train an image encoder that learns to estimate these labels from a crop, similarly to \cite{holdgeoffroy-cvpr-17}. 

In short, our main contributions are the following: 
\begin{itemize}[noitemsep,topsep=0pt]
\item we propose a novel sky autoencoder, dubbed ``SkyNet''\footnote{Luckily, it has not (yet) gained artificial consciousness~\cite{cyberdyne-systems-97}.}, that can accurately represent outdoor HDR lighting in a variety of illumination conditions; 
\item we show how HDR lighting can be estimated from a single image, modeling a much wider range of illumination conditions than previous work; 
\item we capture a new dataset of 206 radiometrically calibrated outdoor HDR 360\textdegree \ panoramas; 
\item we demonstrate, through a series of experiments and a user study, that our approach outperforms the state-of-the-art both qualitatively and quantitatively; 
\item we offer a technique to bridge the gap between our implicit parameters representation and the versatility of low-dimensional parametric sky models. 
\end{itemize}

\begin{figure*}[t]
\centering
\includegraphics[width=0.8\linewidth]{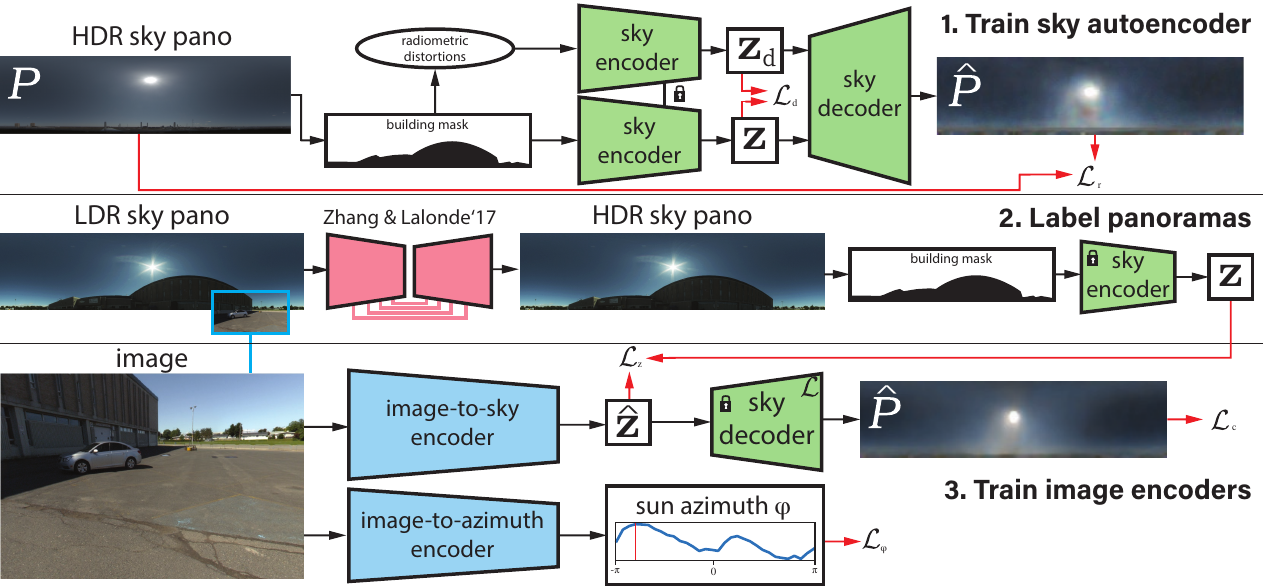}
\vspace{0.15em}
\caption{Overview of our proposed three-step approach. First, we train an autoencoder to learn a 64-parameters latent space of skies~$\mathbf{z}$ from a large dataset of calibrated skies~\cite{hdrdb}, while enforcing its encoder to be robust to distortions in white balance, exposure and occlusions. Second, we convert the SUN360 LDR panorama dataset to HDR using \cite{zhang-iccv-2017} and obtain their $\mathbf{z}$ vectors with the trained autoencoder. Finally, we train two image encoders to learn the mapping between crops from SUN360, the sun azimuth and their corresponding $\mathbf{z}$. Please see text for the definitions of the loss functions $\mathcal{L}_*$. \vspace{-1.5em}}
\label{fig:overview}
\end{figure*}

\section{Related work}

Outdoor lighting modeling and estimation have been studied extensively over the past decades. For conciseness, we will focus on outdoor lighting modeling and estimation that is most related to this work. 

\vspace{-1.3em}
\paragraph{Outdoor lighting modeling} Modeling the sky is a challenging research problem that has been well studied across many disciplines such as atmospheric science, physics, and computer graphics. The Perez All-Weather model~\cite{perez1993allweather} was first introduced as an improvement over the previous CIE Standard Clear Sky model, and modeled weather variations using 5 parameters. Preetham et al.~\cite{preetham-siggraph-99} later present a simplified model, which relies on a single physically grounded parameter, the atmospheric turbidity. Ho\v{s}ek and Wilkie subsequently proposed an improvement over the Preetham model, which is comprised of both a sky dome~\cite{hosek-siggraph-12} and solar disc~\cite{hosek-cga-13} analytical models. See \cite{kider-tog-14} for a comparison of these analytic sky models. 

\vspace{-1.3em}
\paragraph{Outdoor lighting estimation} Lighting estimation from a single, generic outdoor scene has first been proposed by Lalonde et al.~\cite{lalonde-ijcv-12}. Their approach relies on the probabilistic combination of multiple cues (such as cast shadows, shading, and sky appearance variation) extracted individually from the image. Karsch et al.~\cite{karsch2014automatic} propose to match the background image to a large dataset of panoramas~\cite{xiao-cvpr-12} and transfer the panorama lighting (obtained through a specially-designed light classifier) to the image. However, the matching metric may yield results that have inconsistent lighting. Other approaches rely on known geometry~\cite{lombardi2016reflectance} and/or strong priors on geometry and surface reflectance~\cite{barron2015shape}. 

\vspace{-1.3em}
\paragraph{Deep learning for lighting estimation} Deep learning has also been recently used for lighting estimation. For example, Georgoulis et al.~\cite{georgoulis2018reflectance} learn to estimate lighting and reflectance from an object of known geometry, by first estimating its reflectance map (i.e., its ``orientation-dependent'' appearance)~\cite{rematas2016deep} and subsequently factoring it into lighting and material properties~\cite{Georgoulis_2017_ICCV}. Closer to our work, Hold-Geoffroy et al.~\cite{holdgeoffroy-cvpr-17} model outdoor lighting with the parametric, Ho\v{s}ek-Wilkie sky model, and learn to estimate its parameters from a single image. As mentioned above, we take inspiration from this work and significantly improve upon it by proposing to instead use a learned, data-driven outdoor lighting model. Concurrent to this work, Zhang et al.~\cite{zhang-cvpr-19} extend \cite{holdgeoffroy-cvpr-17} with a more flexible parametric sky model. In another closely-related paper, Calian et al.~\cite{calian-eg-18} estimate HDR outdoor lighting from a single face image. While they employ a similar deep autoencoder to learn a data-driven model, they rely on a multi-step non-linear optimization approach over the space of face albedo and sky parameters, which is time-consuming and prone to local minima. In contrast, we learn to estimate lighting from a single image of a generic outdoor scene in an end-to-end framework. In addition, our training procedure is more robust to sky occluders (such as buildings and trees) and non-linear radiometric distortions. Cheng et al.~\cite{cheng2018learning} estimate lighting from the front and back camera of a mobile phone. However, they represent lighting using low-frequency spherical harmonics, which, as shown in \cite{calian-eg-18}, does not appropriately model outdoor lighting.

\section{Overview}
\label{sec:overview}

The goal of our technique is to estimate the illumination conditions from an outdoor image. Directly training such a method in a supervised manner is currently impossible as no large-scale dataset of images and their corresponding illumination conditions is yet available. We therefore propose the following 3-step approach, which is also illustrated in fig.~\ref{fig:overview}. 

\vspace{-1.3em}
\paragraph{1. Train the SkyNet autoencoder on HDR skies} The first step (fig.~\ref{fig:overview}, top row) is to learn a data-driven sky model from the 33,420 hemispherical sky images in the Laval HDR sky database~\cite{hdrdb} using a deep autoencoder. The autoencoder, dubbed ``SkyNet'', learns the space of outdoor lighting by compressing an HDR sky image to a 64-dimensional latent vector $\mathbf{z}$, and reconstructing it at the original resolution. Robustness to white balance, exposure, and occlusions is enforced during training. More details on this step are presented in sec.~\ref{sec:sky-model}.

\vspace{-1.3em}
\paragraph{2. Label LDR panoramas with SkyNet} The second step (fig.~\ref{fig:overview}, middle row) is to use the learned SkyNet autoencoder to obtain $\mathbf{z}$ vectors for a large dataset of panoramas. For this, the Laval HDR sky database cannot be reused as it only contains sky hemispheres. Instead, we take advantage of the wide variety of scenes and lighting conditions captured by the SUN360 panorama dataset~\cite{xiao-cvpr-12}. Each panorama is first converted to HDR with the approach of \cite{zhang-iccv-2017} that has been trained specifically for this purpose. Then, sky masks are estimated using the sky segmentation approach of~\cite{holdgeoffroy-cvpr-17} based on a dense CRF~\cite{krahenbuhl-nips-12}. The resulting HDR panoramas, which we dub SUN360-HDR, along with their sky masks are forwarded to the SkyNet encoder to recover $\mathbf{z}$. This has the effect of labeling each panorama in SUN360 with a compact, data-driven representation for outdoor illumination. 

\vspace{-1.3em}
\paragraph{3. Train image encoders to predict illumination} Finally, the last step (fig.~\ref{fig:overview}, bottom row) is to train an image encoder on limited field of view images extracted from the SUN360 dataset, by employing the methodology proposed in \cite{holdgeoffroy-cvpr-17}. The main difference here is that we train the neural network to predict the $\mathbf{z}$ vector from the previous step corresponding to each crop, instead of the analytical sky parameters as in the previous work. The full HDR sky image can be recovered using the SkyNet decoder. The resulting sky image can be used ``as is'' as image-based lighting to photorealistically render 3D objects into images with a variety of illumination conditions. We detail this step in sec.~\ref{sec:image-head}.

\section{Training the SkyNet deep sky model}
\label{sec:sky-model}

\begin{figure}[t]
\centering
\includegraphics[width=\linewidth]{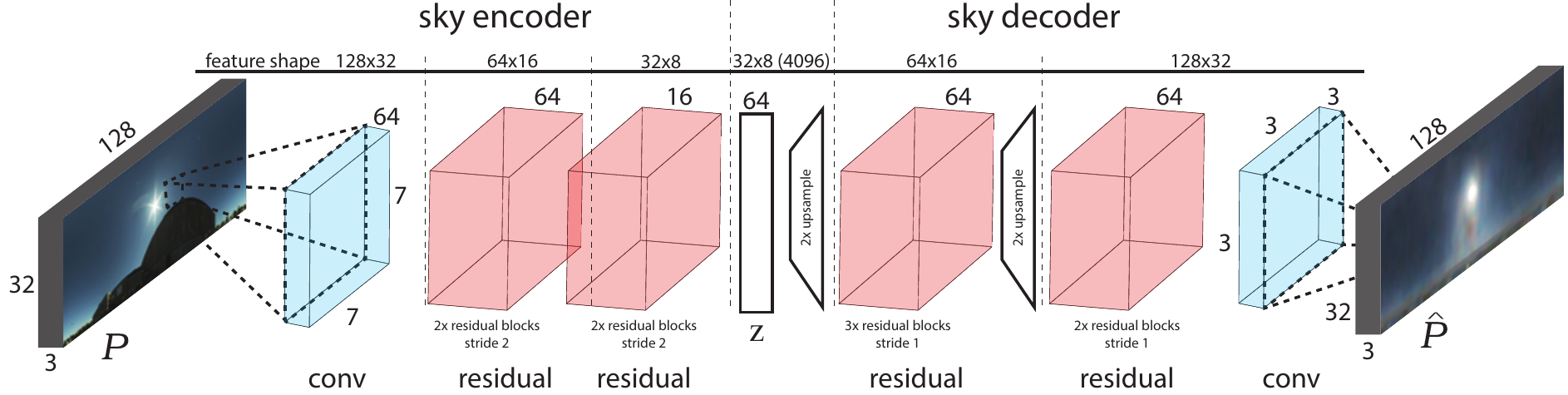}
\caption{Architecture of our SkyNet deep autoencoder showing the parameters of each layer. ELU~\cite{clevert-iclr-16} activation functions are used after the convolutional layers (blue). Residual blocks (red)~\cite{he-cvpr-16} have the ReLU activation functions.}
\label{fig:autoencoder-architecture}
\end{figure}

In this section, we describe SkyNet, our deep autoencoder acting as our sky model, its architecture and training steps. 


\subsection{Deep autoencoder}
\label{sec:autoencoder}

To learn our sky model, we adopt an autoencoder architecture which projects a full HDR sky down to 64 parameters (encoder), and subsequently reconstructs it (decoder). This is conceptually similar to~\cite{calian-eg-18}, with the key differences that we employ a more robust training scheme which includes occlusions and radiometric distortions, making it amenable to full end-to-end learning rather than the non-linear inverse rendering framework of \cite{calian-eg-18}. In addition, we employ a different architecture based on residual layers~\cite{he-cvpr-16} (see fig.~\ref{fig:autoencoder-architecture}). 

To represent the sky, we use the equirectangular (latitude-longitude) projection at a resolution of $32\!\times\!128$ in RGB of the up hemisphere. This representation has the advantage of being easy to rotate along the azimuth with a horizontal shift of the image. Similarly to~\cite{calian-eg-18,zhang-iccv-2017}, we rotate the panoramas along their azimuth so that the sun is in the center of the image, as we empirically found that training the sky model is simpler and more well-behaved this way. However, unlike its azimuth, we cannot decouple the sun elevation from the sky reconstruction as it influences the sun intensity, color, and overall sky luminance distribution~\cite{perez1993allweather}. 

The SkyNet autoencoder training is mostly performed on the 33,420 panoramas of the Laval HDR sky database~\cite{hdrdb,lalonde-3dv-14}, which we augment with 7000 panoramas from SUN360-HDR~\cite{xiao-cvpr-12,zhang-iccv-2017} (see sec.~\ref{sec:overview}), both of which include the full dynamic range of the sun. We resize each panorama down to a resolution of $32\times128$, ensuring that the sky integral remains constant by taking the solid angles into account. 

While the Laval sky database contains unoccluded sky images, panoramas in the SUN360-HDR may contain multiple buildings and other sky occluders which we do not want to learn in our sky model. To prevent SkyNet from learning non-sky features, we reuse the sky segmentation of~\cite{holdgeoffroy-cvpr-17} (based on a CRF refinement~\cite{krahenbuhl-nips-12}) to mask non-sky regions of SUN360-HDR with black pixels. To enforce SkyNet to estimate plausible sky appearance in those regions, we randomly apply black regions to the training images from the Laval sky database and ask the network to recover the original, unoccluded sky appearance. Specifically, we apply, with 50\% chance, the non-sky mask from a random SUN360-HDR panorama. This is only done on the Laval sky panoramas, as SUN360-HDR already contains buildings occluding the sky. This requires the neural network to fill in the holes and predict the sky energy distribution under occluded regions. 

\subsection{Training losses}

\begin{table}[!t]
\centering
\footnotesize
\setlength{\tabcolsep}{4pt}
\begin{tabular}{llll}
\toprule
Parameter & Equation & Distribution & Bounds \\
\midrule
Exposure (e) & $\mathbf{P}_d = e \mathbf{P}$ & $\mathcal{O}(0.2, \sqrt{0.2})$ & $\left[0.1, 10\right]$ \\
White bal. ($\mathbf{w}$) & $\mathbf{P}_{d,c} = \mathbf{w}_c \mathbf{P}_c$ & $\mathcal{N}(0, 0.06)$ & $\left[0.8, 1.2\right]$ \\
Gamma ($\gamma$) & $\mathbf{P}_d = \mathbf{P}^{1/\gamma}$ & $\mathcal{O}(0.0035, \sqrt{0.2})$ & $\left[0.85, 1.2\right]$ \\
\bottomrule
\end{tabular}
\vspace{.25em}
    \caption{Parameters used to generate radiometrically distorted versions $\mathbf{P}_d$ of the panoramas $\mathbf{P}$. Here, $c$ denotes the color channel, $\mathcal{N}(\mu, \sigma^2)$/$\mathcal{O}(\mu, \sigma^2)$ indicate a normal/lognormal distribution.\vspace{-1em}}
\label{tab:parameters-distortion}
\end{table}

\setlength{\abovedisplayskip}{3pt}
\setlength{\belowdisplayskip}{3pt}

To obtain robustness to occlusions and radiometric distortions, we train SkyNet using a combination of two losses, as illustrated in the top part of fig.~\ref{fig:overview}. First, two versions of the panorama are fed through the network, one after the other: the original~$\mathbf{P}$ and a second one to which we applied random radiometric distortions $\mathbf{P}_d$. These random distortions consist of variations in exposure, white balance and camera response function as described in table~\ref{tab:parameters-distortion}. 

Denoting $\text{enc}({\cdot})$ as the encoder, the first loss used to train the sky autoencoder enforces both the undistorted $\mathbf{z}\!=\!\text{enc}({\mathbf{P}})$ and distorted $\mathbf{z}_d\!=\!\text{enc}({\mathbf{P}_d})$ to be as close as possible by minimizing the L1 norm between them: 
\begin{equation}
    \mathcal{L}_d = \lVert \mathbf{z}_d - \mathbf{z} \rVert_1 \,.
\end{equation}
This loss encourages the sky encoder to be robust to radiometric distortions that may be present in the input panoramas. Our second loss is the typical autoencoder reconstruction loss, with the difference that both the undistorted and distorted inputs must reconstruct the original panorama using an L1 loss: 
\begin{equation}
    \mathcal{L}_r = \lVert \hat{\mathbf{P}} - \mathbf{P} \rVert_1 + \lVert \hat{\mathbf{P}}_d - \mathbf{P} \rVert_1 \,,
\end{equation}
where $\hat{\mathbf{P}}\!=\!\text{dec}({\mathbf{z}})$ and $\hat{\mathbf{P}}_d\!=\!\text{dec}({\mathbf{z}_d})$ are the panoramas reconstructed by the decoder $\text{dec}({\cdot})$. The reconstruction loss $\mathcal{L}_r$ is only computed on sky pixels in the original panorama $\mathbf{P}$. For example, this loss is not active for regions masked by buildings in panoramas from SUN360-HDR, as no ground truth sky appearance is known for this region. The autoencoder is never penalized for any output in these regions. On the Laval HDR sky panoramas, this loss is active everywhere, even for randomly masked (black) regions. The target appearance for those regions is the original sky pixels of the panorama before the sky was masked, effectively asking the autoencoder to extrapolate---or fill---the region with plausible sky appearance. 

Our sky autoencoder is trained with: 
\begin{equation}
    \mathcal{L}_s = \mathcal{L}_r + \lambda{}_d \mathcal{L}_d \,,
\end{equation}
where we empirically set $\lambda{}_d\!=\!100$ in order to balance the gradient magnitude between $\mathcal{L}_r$ and $\mathcal{L}_d$ during training. 

Example sky reconstructions on test panoramas are shown in the middle column of fig.~\ref{fig:dataset-excerpt}. While LDR content such as clouds is lost, the reconstructed panoramas $\hat{\mathbf{P}}$ properly model the energy distribution of the sky and are thus able to faithfully reproduce shadow characteristics like contrast and sharpness. In contrast, while the Ho\v{s}ek-Wilkie sky model properly approximates clear skies, it does not generalize to non-clear skies (right-most column in fig.~\ref{fig:dataset-excerpt}).

\subsection{Implementation details} 
\label{sec:sky-model-implementation}

Our sky model holds approximately 1 million parameters which are learned using the Adam~\cite{kingma-iclr-15} optimizer with a learning rate of $10^{-3}$ and $\beta = \left( 0.5, 0.999 \right)$. We additionally reduce the learning rate by a factor of 10 whenever the minimum error on the validation set has not decreased over the last 10 epochs.  Convergence is monitored on the validation set, which is comprised of 14 days (3999 panoramas) from the Laval HDR sky database that we removed from the training set and 2000 panoramas from SUN360-HDR (sec.~\ref{sec:overview}), different from the ones chosen to augment the training set. Training convergence was obtained after 127 epochs in our case, taking roughly 4 hours on a Titan Xp GPU. Sky inference takes approximately 10ms on the same machine.  We (un)normalize the input (output) panoramas using the training set mean and standard deviation. 

\section{Learning to estimate illumination from a single image}
\label{sec:image-head}

\begin{figure}[t]
\centering
\footnotesize
\setlength{\tabcolsep}{2pt}
\begin{tabular}{cc}
\includegraphics[width=0.48\linewidth]{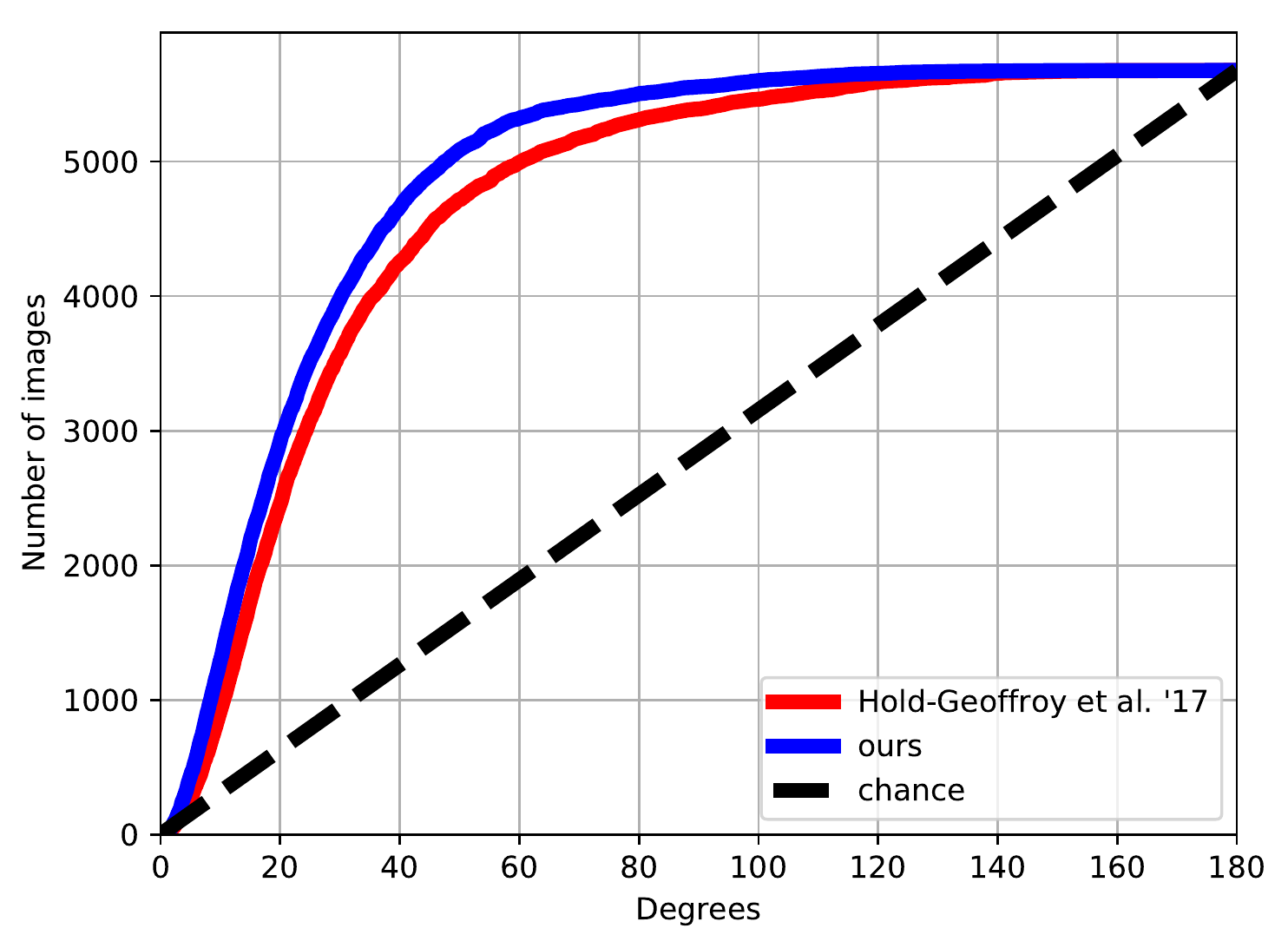} &
\includegraphics[width=0.48\linewidth]{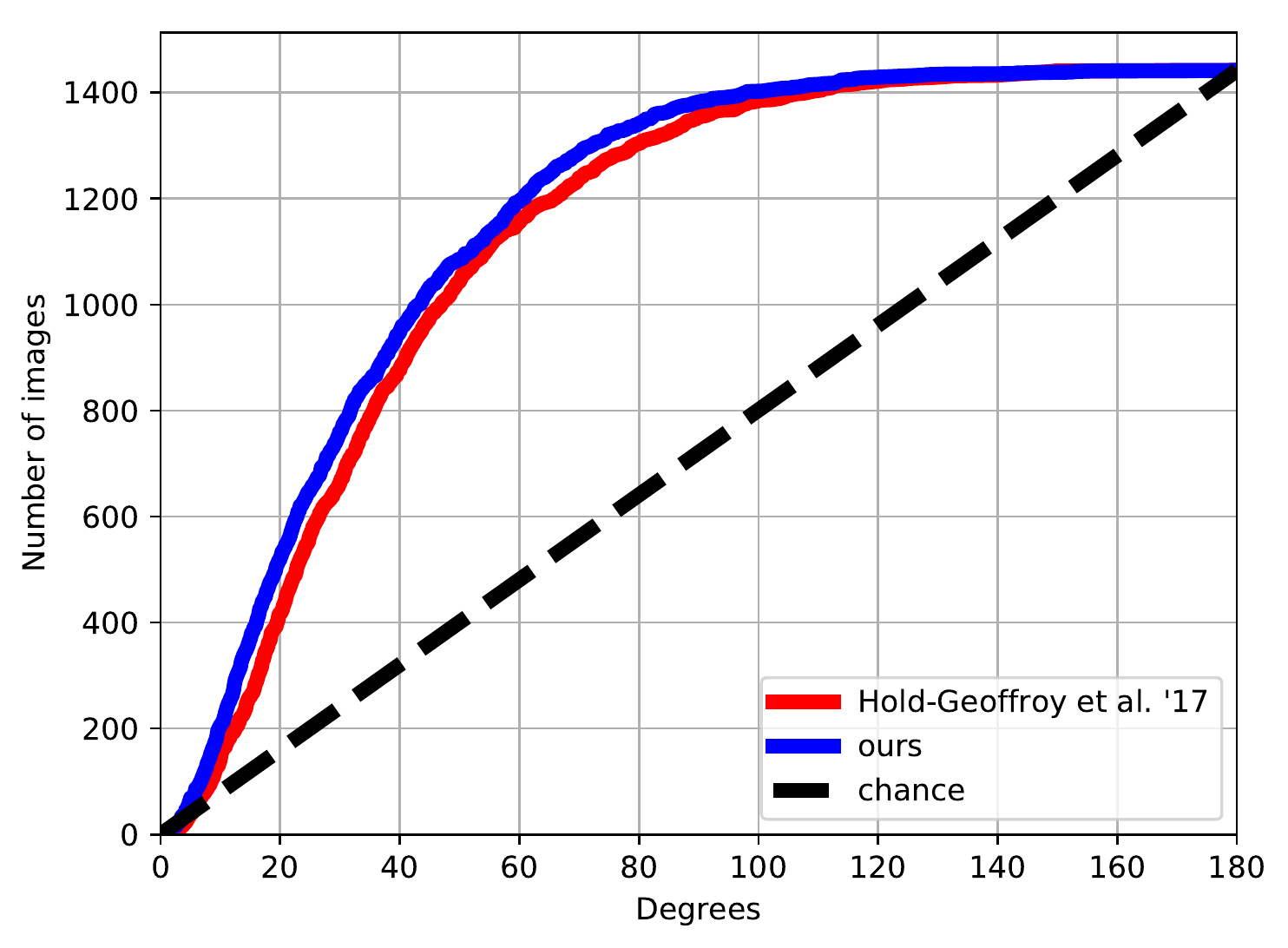} \\
(a) SUN360 & (b) HDR panoramas \\
\end{tabular}
\vspace{.25em}
\caption{Comparison of sun position estimations between our proposed model (blue) and Hold-Geoffroy et al.~\cite{holdgeoffroy-cvpr-17} showing the cumulative sun position estimation error on (a) their SUN360 test set and (b) our HDR 360$^\circ$ captures. Using a recent network architecture (DenseNet-161~\cite{huang-cvpr-16}) grants our technique a slight improvement over the network used by \cite{holdgeoffroy-cvpr-17}. \vspace{-1em}}
\label{fig:results-sun-position}
\end{figure}

In this section, we describe the third step of our approach (c.f. sec.~\ref{sec:overview} and fig.~\ref{fig:overview}), that is, how we learn to estimate both the sun azimuth~$\varphi$ and the sky parameters~$\mathbf{z}$ of our learned sky model from a single, limited field of view image.


\subsection{Image lighting estimation}
\label{sec:image-to-sky}

\setlength{\abovedisplayskip}{3pt}
\setlength{\belowdisplayskip}{3pt}

To estimate the sky parameters~$\mathbf{z}$ from a limited field of view image, we use a pretrained DenseNet-161~\cite{huang-cvpr-16} architecture where the last layer was replaced by a fully connected layer of 64 outputs. We finetune this image-to-sky model on sky parameters~$\mathbf{z}$ using an L2 loss:
\begin{equation}
    \mathcal{L}_z = \lVert \mathbf{\hat{z}} - \mathbf{z} \rVert_2 \,.
\end{equation}
We observed that this loss on the space of $\mathbf{z}$ alone failed to capture the details in the sky energy distribution and tended to produce average skies without strong sun intensities. 
To solve this issue, we added an L1 loss on the sky panoramas reconstructed from $\mathbf{\hat{z}}$ and $\mathbf{z}$ by the SkyNet decoder: 

%
\begin{equation}
    \mathcal{L}_c = \lVert ( \text{dec}({\mathbf{\hat{z}}}) - \text{dec}(\mathbf{z})) \odot \mathbf{d\Omega{}} \rVert_1 \,,
\end{equation}
where $\text{dec}(\cdot)$ denotes the SkyNet decoder, $\mathbf{d\Omega}$ the matrix of solid angles spanned by each pixel in the sky panorama, and $\odot$ the element-wise multiplication operator. 

The image-to-sky encoder is trained by summing those two losses: $\mathcal{L}_i = \mathcal{L}_z + \lambda{}_c \mathcal{L}_c$. Due to the large difference in magnitude between $\mathcal{L}_z$ and $\mathcal{L}_c$, we empirically set $\lambda{}_c = 3 \times 10^{-10}$ to prevent gradient imbalance during training. 

\subsection{Sun azimuth estimation}
\label{sec:image-to-azimuth}

Due to our sky model training (sec.~\ref{sec:autoencoder}), the sun will invariably be located in the center column of the estimated sky panorama. We therefore need to estimate the sun azimuth~$\varphi$ to rotate the lighting according to the sun position in the image. Both tasks seem to be closely related, hinting that both could benefit from joint training~\cite{holdgeoffroy-cvpr-17,zamir-cvpr-18}. However, training a single model to estimate both the sky parameters~$\mathbf{z}$ and sun azimuth~$\varphi$ proved difficult. In our experiments, balancing both tasks using a fixed ratio between the losses failed to obtain good generalization performance for both tasks simultaneously. To circumvent this issue, we train a different image-to-azimuth model to estimate a probability distribution of the sun azimuth~$\varphi$. This sun azimuth distribution is obtained by discretizing the $\left[-\pi, \pi\right]$ range into 32 bins, similar to~\cite{holdgeoffroy-cvpr-17}. We use once again a pretrained DenseNet-161 where the last layer is replaced by a fully connected layer of 32 outputs. A Kullback-Leibler divergence loss $\mathcal{L}_{\varphi}$ with a one-hot target vector is used to train this neural network.

\begin{figure}[t]
\centering
\footnotesize
\floatbox[{\capbeside\thisfloatsetup{capbesideposition={right},capbesidewidth=5cm}}]{figure}[\FBwidth]
{
\setlength{\tabcolsep}{0pt}
\begin{tabular}{r}
\includegraphics[height=2.9cm,width=2.0cm]{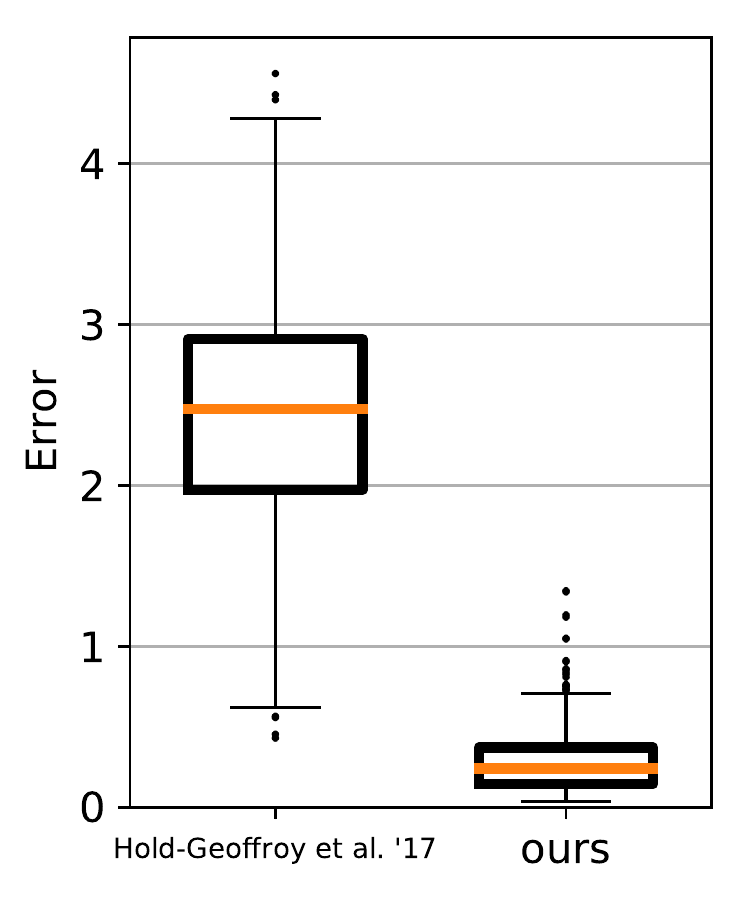} \\
\multicolumn{1}{c}{(a) RMSE} \\
\includegraphics[height=2.9cm,width=2.2cm]{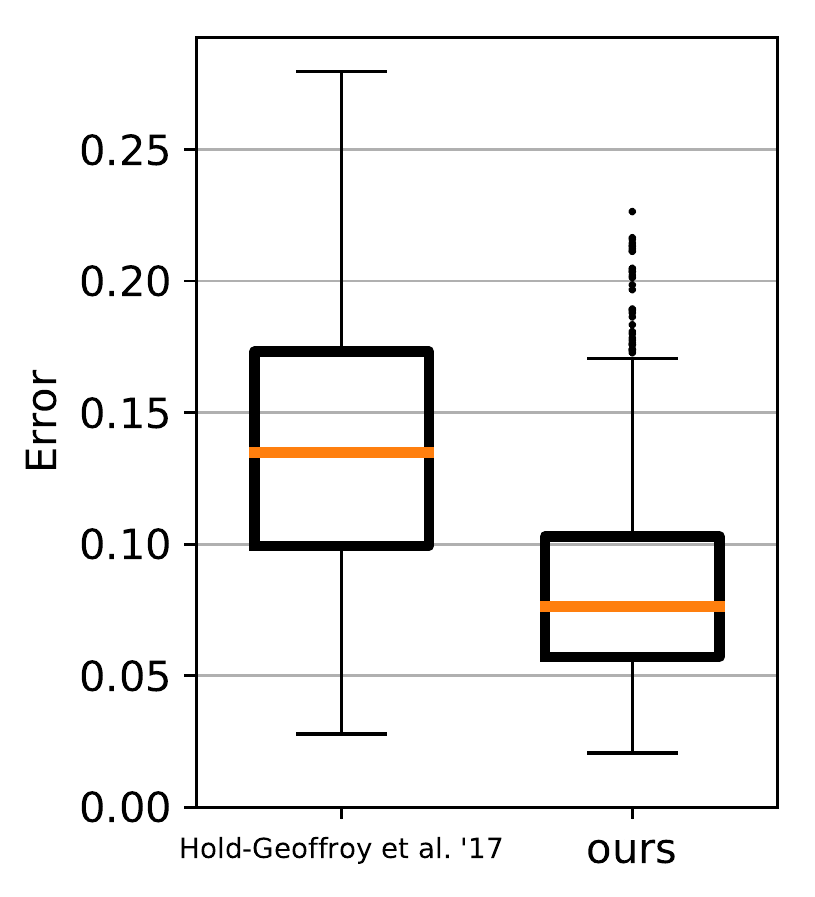} \\
\multicolumn{1}{c}{(b) si-RMSE} \\
\end{tabular}
}{\caption{Quantitative relighting error on the bunny scene (see fig.~\ref{fig:results-lighting-qualitative-ulavaloutdoorhdr}). We compute two metrics comparing renders against ground truth lighting: (a) RMSE and (b) scale-invariant (si-)RMSE~\cite{grosse-iccv-09}. The lighting has been rotated for both methods so the sun is always at the same azimuth. The global intensity of our estimated environment map is generally closer to the ground truth most of the time, leading to an almost $10\!\times$ improvement in RMSE over~\cite{holdgeoffroy-cvpr-17}. Additionally, our flexible learned sky model allows for increased shadow expressiveness and can handle completely overcast skies, enhancing the si-RMSE by over 60\% over the previous state-of-the-art. \vspace{-1em}
\label{fig:results-lighting-quantitative}}
}
\end{figure}

\subsection{Implementation details}
\label{sec:image-head-implementation-details}

To train both the image-to-sky and image-to-azimuth encoders, we use the SUN360-HDR dataset which we augment with 100 images extracted from 15 captured HDR panoramas (see sec.~\ref{sec:ulaval-outdoor-hdr} for more details). To counter the severe imbalance between both data sources, we penalize errors committed on captured panoramas by a factor of 4. 

To provide more stability to the training, the image-to-sky encoder is first trained for 5 epochs using a learning rate of $3\!\times\!10^{-4}$ using only the loss on sky parameters $\mathcal{L}_z$. Afterward, both losses $\mathcal{L}_c$ and $\mathcal{L}_z$ are combined and the learning rate is set to $2\!\times\!10^{-6}$. The image-to-azimuth model was trained with a fixed learning rate of $3\!\times\!10^{-4}$. The Adam optimizer is used with $\beta\!=\!\left(0.4, 0.999\right)$ and a weight decay of $10^{-7}$ throughout the training for both the image-to-sky and sun azimuth estimator. Convergence of the image-to-sky and image-to-azimuth models were obtained after 55 and 3 epochs (roughly 5 hours of training each on a Titan Xp GPU), and inference takes roughly 30ms and 24ms, respectively. 

\begin{figure}[t]
\centering
\footnotesize
\newcolumntype{C}[1]{>{\centering\let\newline\\\arraybackslash\hspace{0pt}}m{#1}}
\setlength{\tabcolsep}{0pt}
\begin{tabular}{C{2.90cm}C{1.75cm}C{1.75cm}C{1.75cm}}
input image & GT & ours & \cite{holdgeoffroy-cvpr-17} \\
\end{tabular}
\includegraphics[width=\linewidth]{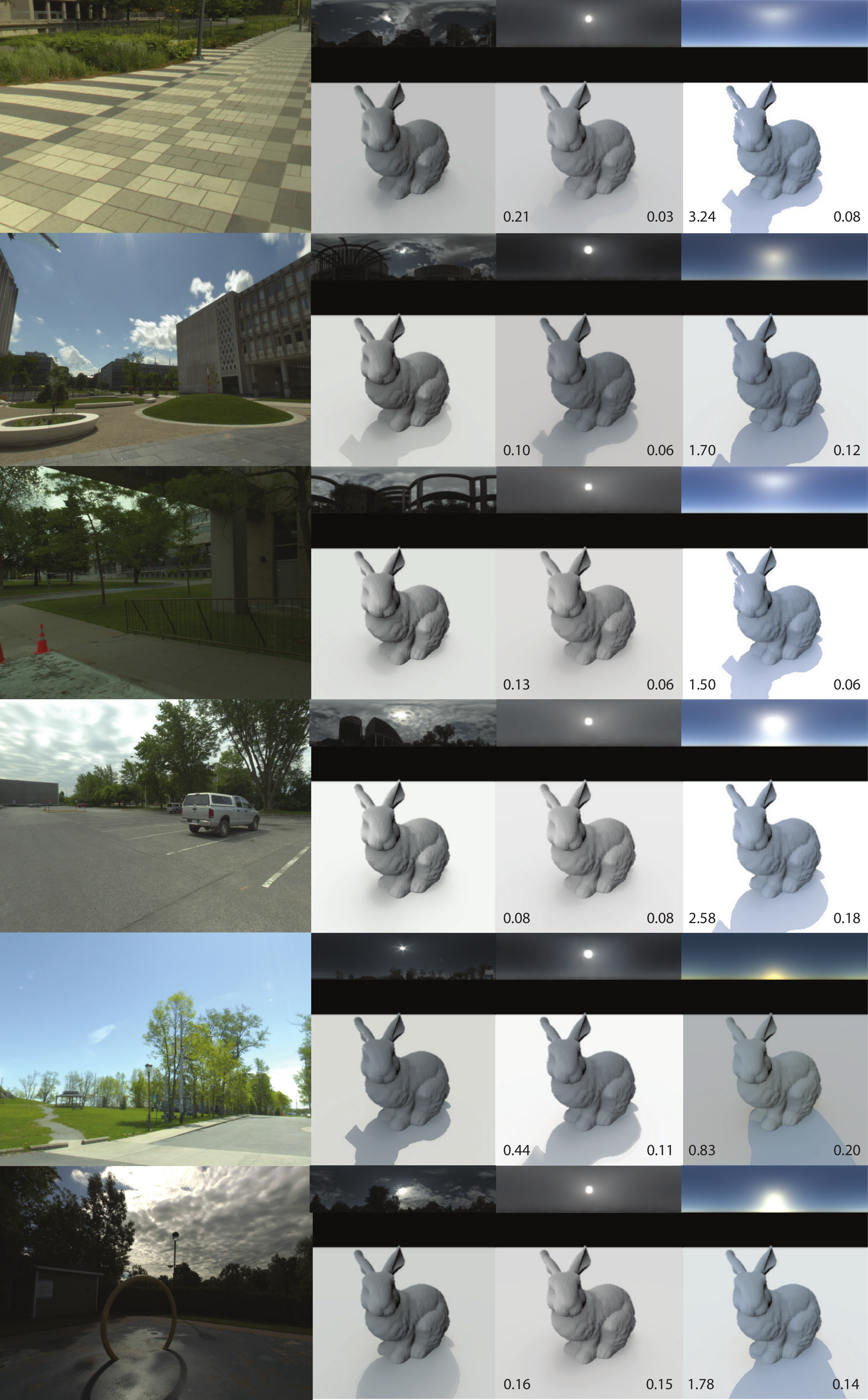}
\caption{Qualitative relighting comparison between ground truth lighting (GT), our method, and Hold-Geoffroy et al.~\cite{holdgeoffroy-cvpr-17}. RMSE (SI-RMSE) are shown on the bottom left (right). Images from our HDR outdoor panorama dataset were cropped to obtain the input image. The renders using our estimated lighting display a wide variety of cast shadow characteristics such as sharp (sunny), smooth (lightly overcast) and absent (mostly overcast), which the parametric sky model of~\cite{holdgeoffroy-cvpr-17} cannot reproduce. }
\label{fig:results-lighting-qualitative-ulavaloutdoorhdr}
\end{figure}

\section{Experimental validation}

This section first presents the dataset used for evaluating and comparing our method to the state-of-the-art method of Hold-Geoffroy et al.~\cite{holdgeoffroy-cvpr-17}. Then, the performance of our proposed method is assessed with qualitative and quantitative results as well as a user study. 

\subsection{A dataset of outdoor HDR panoramas}
\label{sec:ulaval-outdoor-hdr}


The previous state-of-the-art on outdoor illumination estimation~\cite{holdgeoffroy-cvpr-17} proposed an evaluation based solely on SUN360, where the ground truth was obtained using their non-linear optimization on sky pixels to estimate sun intensity. We argue that evaluating on SUN360 does not provide an accurate quantitative relighting performance since it assumes that the non-linear fit accurately models all types of skies present in the panoramas, which is not the case (fig.~\ref{fig:dataset-excerpt}). 

To provide a more accurate assessment of the performance of our technique, we captured a new dataset of 206 HDR outdoor panoramas\footnote{Available at \url{http://outdoor.hdrdb.com}.}. Following the recommendations of~\cite{stumpfel-afrigraph-04}, each panorama captures the full 22 f-stops required to record the full unsaturated dynamic range of outdoor scenes. Using a Canon 5D Mark iii camera with an 8mm Sigma fisheye lens, a ND3.0 filter, and mounted on a GigaPan tripod head, we captured 6 sets (at 60\textdegree \ azimuthal increments) of 7 exposures (from 1/8000s to 8s shutter speed at f/14 aperture) in RAW mode. We then automatically stitched the results into a 360\textdegree \ HDR panorama using the PTGui commercial software. Since capturing the necessary 42 photos required approximately 3 minutes, care was taken to select scenes with no motion. We repeated this process over 9 different days to capture a diverse set of scenes, resulting in a mix of urban and natural scenes with illumination conditions ranging from overcast to sunny. We select 191 panoramas from this set (non-overlapping in both location and illumination conditions with the 15 used for training in the image-to-sky encoder, see sec.~\ref{sec:image-head-implementation-details}) and extract 7 crops per panorama for a total of 1,337 images, which we use for evaluation below. 

\subsection{Quantitative sun position evaluation}
\label{sec:results-sun}

\begin{table}[!t]
\centering
\footnotesize
\begin{tabular}{rcc}
\toprule
& ours & \cite{holdgeoffroy-cvpr-17} \\
\midrule
total votes & 536 (69\%) & 244 (31\%) \\
\bottomrule
\end{tabular}
\vspace{.15em}
\caption{Results of our user study ($N=39$), which show that users overwhelmingly prefer results obtained with our technique over that of Hold-Geoffroy et al~\cite{holdgeoffroy-cvpr-17}. \vspace{-1em}}
\label{tab:userstudy}
\end{table}

We begin by evaluating the performance of our models in estimating the relative position of the sun with respect to the camera from a single limited field of view image. Results on both the SUN360 test set from~\cite{holdgeoffroy-cvpr-17} (left) and our panorama dataset (right) are shown in fig.~\ref{fig:results-sun-position}. In both cases, the ground truth is obtained by detecting the center of mass of the brightest region in the panorama, following~\cite{holdgeoffroy-cvpr-17} (who reported a median error of $4.59^\circ$). Since our method only estimates explicitly the sun azimuth, the elevation angle is estimated as the brightest pixel of the reconstructed lighting panorama. Due to the more advanced network architecture employed, we systematically improve sun position estimation over the previous state-of-the-art on both datasets.

\subsection{Lighting evaluation on HDR panoramas}
\label{sec:results-lighting}

The relighting error is compared between~\cite{holdgeoffroy-cvpr-17} and our method using the bunny scene on the 1,337 images from our dataset with ground truth illumination (sec.~\ref{sec:ulaval-outdoor-hdr}). Both the RMSE and the scale-invariant (si-)RMSE~\cite{grosse-iccv-09} are computed, and results are shown in fig.~\ref{fig:results-lighting-quantitative}. Our technique yields significant improvement in both the RMSE and si-RMSE. For the RMSE, the improvement is mostly due to the fact that the exposure estimation of~\cite{holdgeoffroy-cvpr-17} that seems biased toward bright skies. The render intensity using our estimated lighting is generally much closer to the ground truth. Additionally, the increased versatility of our sky model confers an additional 60\% improvement on scale-invariant RMSE. 

\begin{figure}[t]
\centering
\includegraphics[width=0.9\linewidth]{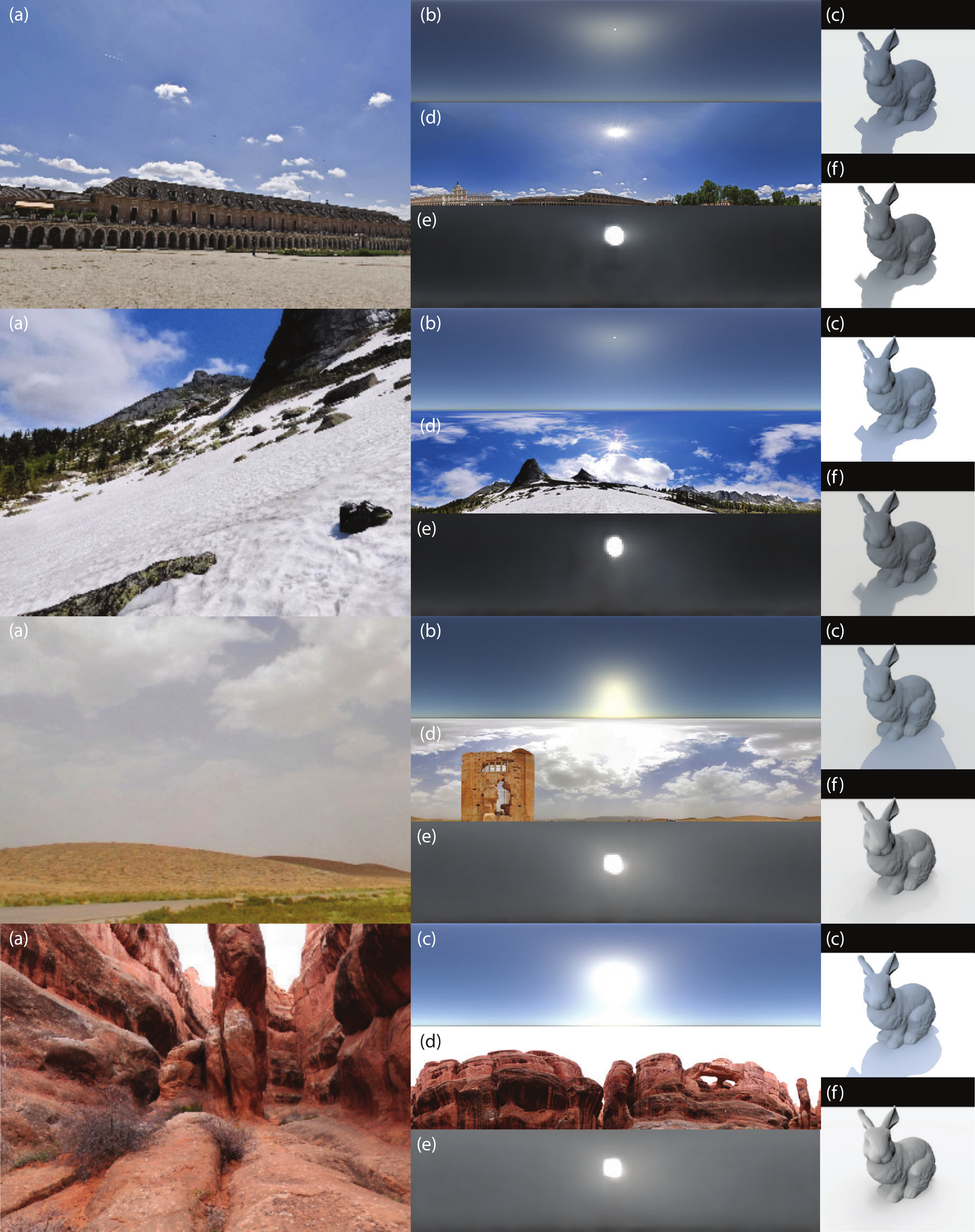} \\
\caption{Qualitative relighting evaluation. From (a) an input image, we show the lighting estimation and render from Hold-Geoffroy et al.~\cite{holdgeoffroy-cvpr-17} (b-c) and our method (e-f) on SUN360. Note that no ground truth illumination exists for this dataset, only (d) a saturated LDR panorama. Our method confers a wider variety of shadow characteristics (f) over that of \cite{holdgeoffroy-cvpr-17}.
\vspace{-.5em}
}
\label{fig:results-lighting-qualitative-sun360}
\end{figure}

Qualitative examples of recovered illumination and renders are shown for test images in our HDR panorama dataset in fig.~\ref{fig:results-lighting-qualitative-ulavaloutdoorhdr} and SUN360 dataset in fig.~\ref{fig:results-lighting-qualitative-sun360}. Both techniques provide plausible estimates yielding strong shadows on sunny days. For both datasets, we observe that lighting from~\cite{holdgeoffroy-cvpr-17} is consistently brighter than the ground truth, resulting in strong cast shadows even on partially cloudy and overcast skies. Our sky model captures the subtle lighting distribution in these conditions more accurately.

\begin{figure}[t]
\centering
\footnotesize
\newcolumntype{C}[1]{>{\centering\let\newline\\\arraybackslash\hspace{0pt}}m{#1}}
\setlength{\tabcolsep}{0pt}
\begin{tabular}{C{2.20cm}C{2.30cm}C{2.10cm}}
Ground Truth & \cite{holdgeoffroy-cvpr-17} & ours \\
\end{tabular}
\includegraphics[width=0.8\linewidth]{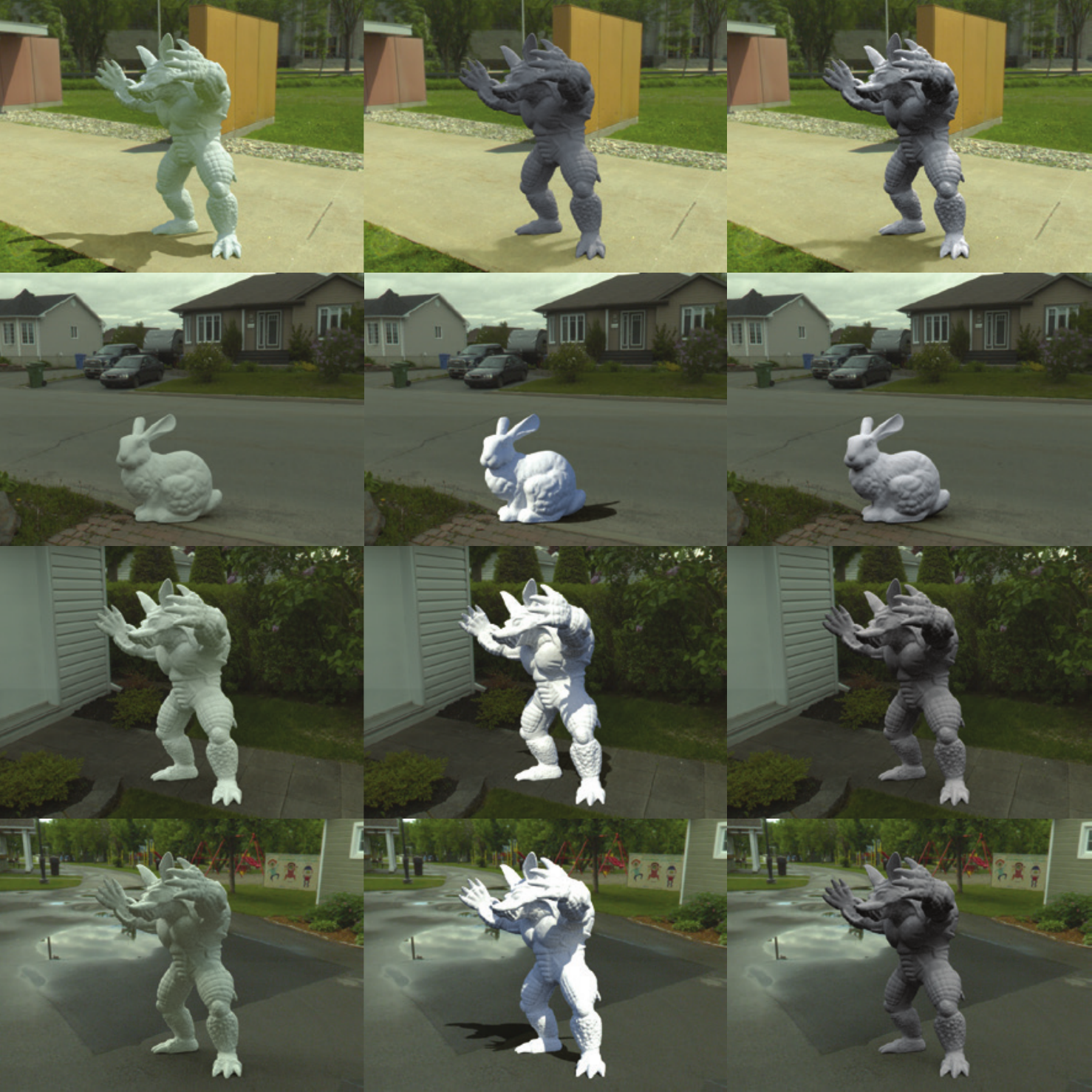}
\caption{Examples of virtual object insertion comparing our method to~\cite{holdgeoffroy-cvpr-17} on backgrounds extracted from our evaluation dataset (see sec.~\ref{sec:ulaval-outdoor-hdr}). \vspace{-1em}}
\label{fig:renders}
\end{figure}

We further compare our method by performing virtual object insertions using the Cycles renderer. As fig.~\ref{fig:renders} shows, our estimated overall lighting intensity is closer to the ground truth than \cite{holdgeoffroy-cvpr-17} while still providing plausible shading. 

\subsection{User study on SUN360 LDR panoramas}
\label{sec:results-userstudy}

As no accurate ground truth illumination is available for SUN360, no quantitative relighting evaluation can faithfully be performed on this dataset. Instead, we evaluate performance with a user study where we showed ($N=39$) participants 20 pairs of images with a virtual object (a bird statue model) composited into the image and lit by the estimates provided by our method and \cite{holdgeoffroy-cvpr-17}. For each pair, users were asked to select the image where the object looked most realistic. 
As shown in tab.~\ref{tab:userstudy}, our method obtained slightly more than 68\% of the total votes. Furthermore, our lighting estimations were preferred (more than 50\% votes) on 16 of the 20 images. 


\section{User-guided edits on the sky model}

\begin{figure}[t]
\centering
\includegraphics[width=\linewidth]{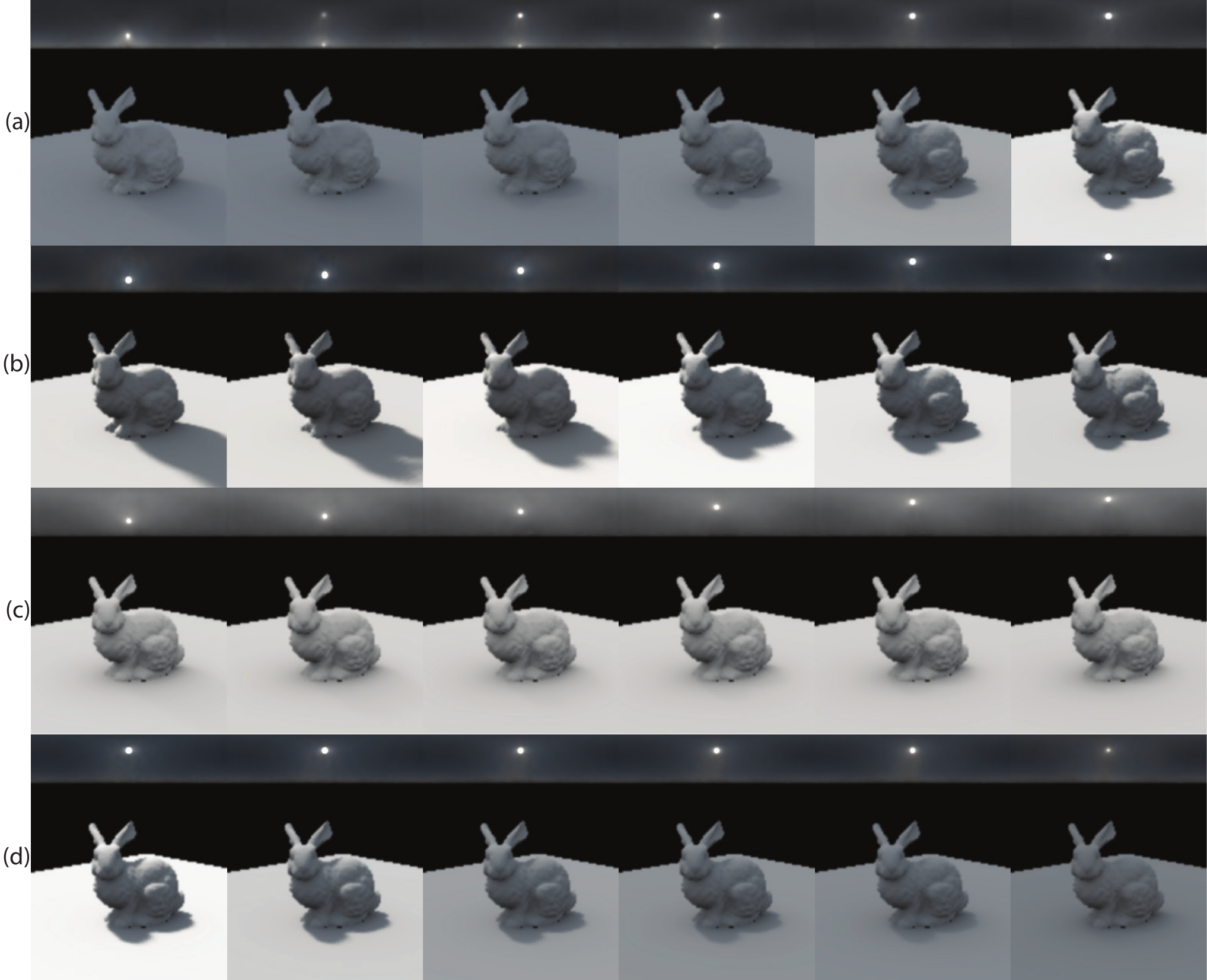}
\caption{Examples of user-guided edits on the sky model. (a) Interpolating between two sky parameters~$\mathbf{z}$ does not produce a smooth and plausible lighting transition. To solve this, we propose a method to enable smooth user-guided lighting edits and show results on changing the sun position on (b) a cloudy and (c) a clear day. Note how the generated skies stay plausible throughout the transition. We further show an example of changing the sun intensity (d), from fully visible to mostly occluded. \vspace{-1em}}
\label{fig:analysis-knobs}
\end{figure}

Low-dimensional analytical sky models such as~\cite{hosek-cga-13,lalonde-3dv-14} provide explicit parameters for users to interact with, such as sun position and atmospheric turbidity. The main drawback of our implicit sky parameters~$\mathbf{z}$ is that they cannot be directly hand-tuned. One could think of generating new skies by interpolating between two sky parameters~$\mathbf{z}$. However, doing so yields abrupt transitions that often contain undesirable artifacts such as multiple suns (fig.~\ref{fig:analysis-knobs}-a). 

We propose a new method to browse the parameter space spanned by~$\mathbf{z}$ while producing smooth transitions and plausible skies. Our intuition is to start from a known sky parameter~$\mathbf{z}$ and iteratively morph this sky toward a desired target using the sky decoder gradient. To generate this gradient, a sky is first forwarded through the sky autoencoder. Then, edits are applied to the reconstructed sky and the resulting gradients on $\mathbf{z}$ are computed. We experiment on two types of edits: the sun elevation and intensity. To change the sun elevation, we move the $5\!\times\!5$ region around the sun either up or down and compute the gradient using the difference between the reconstruction and this modified sky. A similar scheme is used for sun intensity, where the region is multiplied such that its maximum value is the desired sun intensity. Iterating on this scheme using $\mathbf{z}_{n+1} = 4\!\times\!10^{-10} \cdot \frac{\partial{\mathcal{L}_r}}{\partial{\mathbf{z}}} \cdot \mathbf{z}_{n}$ for a maximum of 300 iterations automatically re-projects this modified sky back to a plausible sky and successfully removes the manually-induced artifacts. The multiplying factor was empirically set as a good balance between stability and convergence speed. Visually smooth transitions are shown in fig.~\ref{fig:analysis-knobs}.


\section{Discussion}

In this paper, we propose what we believe is the first learned sky model trained end-to-end and show how to use it to estimate outdoor lighting from a single limited field of view images. Our key idea is to use three different datasets in synergy: SUN360~\cite{xiao-cvpr-12}, Laval HDR sky database~\cite{hdrdb}, and our own HDR 360$^\circ$ captures. Through quantitative and qualitative experiments, we show that our technique significantly outperforms the previous state-of-the-art on both lighting reconstruction and estimation.

While our method proposes state-of-the-art performance, it suffers from some limitations. Notably, the Ho\v{s}ek-Wilkie model employed by~\cite{holdgeoffroy-cvpr-17} tends to produce stronger lighting and sharper shadows than our model, which users seemed to prefer sometimes in our study. 
Additionally, while our model accurately captures the sky energy, its texture recovery quality is still limited. We hope these limitations can be soon lifted by the current rapid development of deep learning.



\section*{Acknowledgements}

The authors wish to thank S\'{e}bastien Poitras for his help in collecting the HDR panorama dataset. This work was partially supported by the REPARTI Strategic Network and the NSERC Discovery Grant RGPIN-2014-05314. We gratefully acknowledge the support of Nvidia with the donation of the GPUs used for this research, as well as Adobe for generous gift funding.

{\small
\bibliographystyle{ieee}
\bibliography{refs}
}

\end{document}